\documentclass[a4paper]{article}
\usepackage[utf8]{inputenc}
\usepackage[T1]{fontenc}
\usepackage{fullpage}
\usepackage{comment}
\usepackage{amssymb}
\usepackage{hyperref} 
\usepackage{amsmath,bm,color}
\usepackage{amsthm}

\usepackage{algorithm}
\usepackage{algorithmic}
\usepackage{xcolor}
\usepackage{mathtools}
\usepackage{graphicx}
\usepackage{subcaption}
\usepackage{authblk}
\usepackage{upref}
\usepackage[toc,page]{appendix}
\usepackage[backend=biber,
    style=ieee,sorting=none]{biblatex}
\definecolor{purple}{rgb}{.4,0,.4}
\definecolor{grey}{rgb}{0.5, 0.50, 0.5}

\usepackage{booktabs} 
\usepackage{multirow}

\AtBeginDocument{ 
  \label{CorrectFirstPageLabel} 
} 

\author[1*]{Pål V. Johnsen}
\author[1]{Eivind Bøhn}
\author[1]{Sølve Eidnes}
\author[1]{Filippo Remonato}
\author[1]{Signe Riemer-Sørensen}

\affil[1]{SINTEF Digital, Oslo, Norway}
\affil[*]{Corresponding author: pal.johnsen@sintef.no}

\title{Recency-Weighted Temporally-Segmented Ensemble for Time-Series Modeling \\
    \large Multi-step Forecasting in Process Industries}

\begin{document}
\maketitle

\begin{abstract}
Time-series modeling in process industries faces the challenge of dealing with complex, multi-faceted, and evolving data characteristics. Conventional single model approaches often struggle to capture the interplay of diverse dynamics, resulting in suboptimal forecasts. Addressing this, we introduce the Recency-Weighted Temporally-Segmented (ReWTS, pronounced `roots') ensemble model, a novel chunk-based approach for multi-step forecasting. The key characteristics of the ReWTS model are twofold: 1) It facilitates specialization of models into different dynamics by segmenting the training data into `chunks' of data and training one model per chunk. 2) During inference, an optimization procedure assesses each model on the recent past and selects the active models, such that the appropriate mixture of previously learned dynamics can be recalled to forecast the future. This method not only captures the nuances of each period, but also adapts more effectively to changes over time compared to conventional `global' models trained on all data in one go. We present a comparative analysis, utilizing two years of data from a wastewater treatment plant and a drinking water treatment plant in Norway, demonstrating the ReWTS ensemble's superiority. It consistently outperforms the global model in terms of mean squared forecasting error across various model architectures by 10-70\% on both datasets, notably exhibiting greater resilience to outliers. This approach shows promise in developing automatic, adaptable forecasting models for decision-making and control systems in process industries and other complex systems.

\end{abstract}

\section{Introduction}

Process industries are concerned with the processing of raw materials into products. In order to achieve the desired quality, this processing must be controlled based on the conditions in which it takes place. For example, for a chemical process to run in an optimal way, the chemical reactions can be controlled through chemical dosing.

The processing system is often complex, and consists of several interconnected stages. The system is also often regulated by so-called closed-loop controllers that, given feedback from the system, make decisions in order to control the state of the system \cite{bequette_process_2010}. An example of a closed-loop controller is the proportional-integral-derivative (PID) controller \cite{PID_future_2001}. However, this type of control is solely reacting based on the present situation, and does not support incorporating future information such as forecasts. An alternative control technique called model predictive control (MPC) is applicable if one can account sufficiently for future consequences of different actions. To employ MPC, a requirement is to have a \textit{forecasting model} that can predict the state of the system in the near future as a function of the control inputs with sufficient accuracy \cite{schwenzer_review_2021}. 
The forecasting model can be entirely based on known physical or chemical system properties, which entails that certain assumptions need to be made. It can also be entirely data-driven or model-free based on historic measurements, with no inherited information about known causal, physical or chemical properties in the system \cite{nauman_model-free_2022}. A hybrid model is a mix of the two types, combining both data-driven methods, as well as scientific and causal knowledge about the process.
 
Causal knowledge can come in many forms. It can be the awareness of intentional changes such as maintenance of equipment, often carried out at regular time intervals. One example of this is the flushing of filters in the drinking water treatment plant considered in Section \ref{sec:bergen}. The process can also be affected by external factors such as weather conditions, and other external factors such as seasons may introduce periodic patterns in the process. At the same time, the process relies on situation-specific occurrences that vary stochastically, or can be intractable to determine in advance. This can for instance be clogging of pipes or sudden weather changes. 

The corresponding stochastic processes of the variables governing the industrial process do not typically exhibit \textit{stationarity} (constant mean and covariance over time). Trends and seasonalities are often seen in the time series, but not necessarily in a consistent, deterministic way, rendering transformations to stationarity not straightforward.
An additional complicating factor is the fact that often many variables in the process are statistically correlated. 
Consequently, we deal with a \textit{multivariate}, \textit{correlated}, \textit{non-stationary}, and partly \textit{seasonal} time series. In this context, it is crucial that the forecasting model can adapt to new patterns, while also being able to recognize previously seen patterns. In this work we present a strategy for reaching this goal, in particular for data-driven forecasting models. The idea is to construct models for disjoint historical regions, and combine them in such a way that previously seen patterns in the data, implicitly learned by the models, can be captured efficiently at prediction time. 

In Section \ref{sec:problem} we formulate the problem, introduce notation and list some assumptions. In Section \ref{sec:related} we discuss proposed solutions from the literature and explain why they are insufficient. Our novel method is given in Section \ref{sec:Chunk}, before we test it on simulated and real data in Section \ref{sec:Results}. Section \ref{sec:Discussions} provides a discussion of the strengths and weaknesses of the method and concluding remarks.

\section{Problem description, assumptions and notation} \label{sec:problem}
Given multivariate time series data describing a certain process, we consider the aim of forecasting the future behaviour of the target variable $y$. For each time point, along with a measurement from $y$, there are also measurements from $K$ variables, from hereon denoted covariates. For simplicity, assume the target and the $K$ covariate measurements are consistently available or otherwise estimated at equidistributed time points. Let $t_0 = 0$ be the first available recording time of all measurements, while $t_n$ denotes the present time point. The time interval between two consecutive time points is denoted $\Delta t$. We let $\mathbf{y}_{a:b}$ be the target vector of size $b-a+1$, and $X_{a:b}$ the covariate matrix of dimension $(b-a+1)\times K$ that include all the measurements between $t = t_a$ and $t = t_b$. At any prediction time, $t_n$, we want to construct a \textit{forecasting model} that can predict the target measurements between $t= t_{n+1}$ and $t = t_{n+h}$, where $h$ denotes the \textit{forecast horizon}.

Let the vector $\hat{\mathbf{y}}_{(n+1):(n+h)}$ denote the $h$ predicted future measurements at time $t_n$ of the target variable from a particular forecasting model as a function of historic data of both the target and the covariates. In some cases, the future values of the $K$ covariates up to time $t_{n+h}$, included in the matrix $X_{(n+1):(n+h)}$, are known. An example relevant in process industries is where a covariate represents planned operations or control signals such as chemical dosing, or when filters are flushed. In this case, the forecasts of the target can also be a function of these already-known future covariate values.

Assume we have historic data available including known values of the target and a corresponding trained forecasting model. We want to evaluate the performance of the forecasting model with respect to the historic data. At prediction time, $t = t_n$, the performance of the forecasting model given particular forecast values $\hat{\mathbf{y}}_{(n+1):(n+h)}$ can be quantified by a similarity measure, also called a \textit{loss function}, $L(\mathbf{y},\mathbf{\hat{y}})$. For simplicity we will in this work focus on the mean squared error (MSE) as the loss function:

\[L\left(\mathbf{y}_{(n+1):(n+h)},\hat{\mathbf{y}}_{(n+1):(n+h)}\right) = \frac{1}{h}\sum_{i=1}^h (y_{n+i}-\hat{y}_{n+i})^2.\]

Given all historic data from $t_f$ up to time $t_e$, denoted $X_{f:e}$ and $\mathbf{y}_{f:e}$, we want to evaluate the $h$-step forecasts from the forecasting model not only at a single prediction time point, but by considering its predictions on the entire span of predictions between $t = t_f$ to $t = t_e$. This can be achieved through averaging all forecast loss values for individual time stamps from $t=t_f$ to $t=t_e$ with a \textit{stride} $s$ (the number of time steps between successive forecasts).

\begin{equation}
\resizebox{0.97\textwidth}{!}{$L_s(\mathbf{y}_{f:e},\mathbf{\hat{y}}_{(f+1):(f+h)},\ldots,\mathbf{\hat{\mathbf{y}}}_{(f+\psi s + 1):(f+\psi s +h)}) = \frac{1}{\psi}\sum_{k=0}^{\psi} L\left(\mathbf{y}_{(f + ks+1):(f+ks+h)},\hat{\mathbf{y}}_{(f + ks+1):(f + ks+h)}\right),$}
\label{loss}
\end{equation}

with $\psi = \lfloor (e-h-f)/s \rfloor$.

\section{Related work} \label{sec:related}

The challenge of continuously training a machine learning model with new incoming data streams is described in \cite{mccloskey_catastrophic_1989}, where the term \textit{catastrophic forgetting} is introduced. This describes the situation where a model by learning relationships from new data, undesirably forgets previously learned relationships. The consequence is that even though it would be intuitive and practical to update a machine learning model trained on historic data with a batch of the most recent data, it may cause the model to forget previously observed relationships. The alternative is to retrain the model from scratch, including both old and new data.

An alternative to a single model that needs tuning or complete retraining for new incoming data, is to have several models trained from different time intervals, and combine them in an appropriate way to give the final prediction. The idea is that the different models are trained on different dynamics, so that each model is specialized on a certain subset of the global dynamics. At forecasting time, the present dynamics might be a combination of several dynamics previously seen by the models, such that a specific weighted combination or \textit{ensemble} of the models may give a reasonable forecast. 

A comprehensive survey of ensemble learning with data streams is given in \cite{krawczyk_ensemble_2017}. In this survey, the authors distinguish between \textit{online learning} and \textit{chunk-based learning}. Online learning approaches process each new incoming sample sequentially, for instance by updating the model at every incoming sample. In chunk-based learning, incoming data is processed in chunks, for instance by fitting a separate model on each chunk. It is further distinguished between \textit{stationary} and \textit{non-stationary} data streams. In the second case, the data generating process is not stable in terms of constant mean and covariance, but in fact changing over time. We say the data exhibits \textit{concept drift} in the form of gradual changes in the data generating process. This is a typical example for process industries.

A typical framework for chunk-based models is to first define a chunk length, adaptively train a model for every chunk of data, and finally combine all models into an ensemble model to predict for new data. Exactly how to combine all chunk models into one ensemble model does not have a definitive answer, and there exist several proposals for this. One proposal, described in \cite{wang_mining_2003} for classification models, is to assign weights to each classifier based on its performance on the newest training chunk. The assumption is that the last training chunk is the most representative of the new incoming data. Here, the weight for each chunk model is computed based on a classification-based mean squared error on the most recent training chunk.

There are also several approaches dealing with how to discard previous models if they are under-performing. E.g., in \cite{wang_mining_2003}, classification models are discarded if they are performing worse than a random classification model. However, as noted in \cite{mccloskey_catastrophic_1989}, this pruning strategy might be too strict in the situation of sudden concept drifts, as is typical for process industry data. 

In the realm of non-stationary data stream analysis, numerous ensemble techniques have been developed to address classification tasks \cite{AUE,learn_plus_plus,SEA}. The exploration of ensemble methods for multi-step forecasting has been less explored at the time of writing. This is in particular true for non-stationary data streams. In \cite{galicia_multi-step_2019}, a heterogeneous ensemble model for multi-step forecasting is presented, however not presented explicitly in the context of continuous data streams, but with evaluation on a static future test data. Moreover, this is not a chunk-based method, but rather only the most recent data of a fixed size is used to construct the ensemble model at prediction time. The ensemble model is a weighted sum of a decision tree model, a gradient boosting model and a random forest model trained on the most recent data, denoted as the training set. The weights are calculated such as to minimized the mean squared error of the forecasts on the last portion of the training set, denoted the validation set. The authors further introduce an updating period for when to recalculate the weights based on the most recent historic data at prediction time.

\section{The Recency-Weighted Temporally-Segmented ensemble} \label{sec:Chunk}
Here we consider a chunk-based ensemble model to be a model consisting of multiple models (be it domain-based, data-driven or a combination) that are combined in some way that involves weighting of the models, to become a single model. The models in the ensemble may be of the same type (homogeneous ensembles), or of different types (heterogeneous ensembles). An example of a homogeneous ensemble is a random forest, where all the models are decision trees. For the chunk-based ensemble model we propose here in the context of time series forecasting, each single model is of the same type, but trained on different time intervals. In the following, we consider only the most naive approach for choosing these intervals, which is to split the timeline in disjoint equidistant sets. However, we note that varying length or overlapping intervals could be other alternatives.

There are several ways the models in an ensemble can be weighted, but here we suggest a linear combination scheme designed to overcome practical challenges in the process industry. Let the matrix $Z_{n}$ characterize the dynamics of the data at time $t_n$, given by recent historic data from a specified historic time point up to the present time point $t_n$. Assume we have the set $\mathcal{M}_{n}$ of already trained forecasting models at time $t_n$, and let the matrix $M_h(X_{:n},\mathbf{y}_{:n})$ of dimension $h \times \lvert\mathcal{M}_{n}\rvert$ include the forecasts for the $h$ following time steps for each model. Then, the $h$-step ahead expected forecast can be written as

\begin{equation*}
E_{\mathcal{M}}[\hat{\mathbf{y}}_{(n+1):(n+h)}] = M_h(X_{:n},\mathbf{y}_{:n})\mathbf{p}_{\mathcal{M}_{n}}(\mathbf{Z}_{n}),
\end{equation*}
where the vector $\mathbf{p}_{\mathcal{M}_{n}}(\mathbf{Z}_{n})$ of size $\lvert\mathcal{M}_{n}\rvert$ can be interpreted as containing the probabilities that each model $m_j \in \mathcal{M}_{n}$ should be used in the ensemble given the dynamics in the data described by $\mathbf{Z}_{n}$. 
From hereon we will denote $\mathbf{p}_{\mathcal{M}_{n}}(\mathbf{Z}_{n})$ as the \textit{weights} $\mathbf{w}(t_n)$ at time $t_n$. Interpreting the weights as probabilities implies the constraints

\begin{align*}
    &w_j(t_n) \ge 0 \quad \forall  \ j,n,  \\
    &\sum_{j=1}^{\mathcal{M}_{t_n}} w_j(t_n) = 1.
\end{align*}

The question then becomes in what way the weights should be estimated at every time point $t_n$, to support the goal of providing robust forecasts. A particular characterization of the dynamics in the data, $\mathbf{Z}_{n}$, is required for this purpose and will set the stage for how the weights are calculated. Below, we suggest one weight assignment procedure by letting $\mathbf{Z}_{n}$ include the most recent historic data from a predefined historic time point and up to time $t_n$. We call this the look-back data.

\subsection{The near past as an indication of the near future}
Based on an idea from \cite{wang_mining_2003} for classification problems, the weights of the models at prediction time $t_n$ can be estimated by looking at the dynamics of the data in the near past of time $t_n$. Intuitively, the best performing weight assignment of the models on recent data is likely also applicable for data in the near future. From the present time point $t_n$, let $\mathbf{Z}_{n}=(X_{(n-l_b):n},\mathbf{y}_{(n-l_b):n})$ denote the look-back data going $l_b$ time steps back in time. We seek the linear combination of previously trained models that minimizes the mean squared error of the $h$ step ahead forecasts across the look-back data. Translating this to a mathematical optimization problem, we search for the weights $w_j(t_n)$, for $j = 1,\ldots,\lvert\mathcal{M}_{n}\rvert$, at time $t_n$ for the ensemble of the models in $\mathcal{M}_{n}$ that solve the following optimization problem:

\begin{equation}
    \begin{aligned}
       &\arg\min_{\mathbf{w}(t_n)} \sum_{k = n-l_b}^{n-h} \Big(\mathbf{y}_{(k+1):(k+h)}-M_h\left(X_{:k},\mathbf{y}_{:k}\right) \mathbf{w}(t_n)\Big)^T \Big(\mathbf{y}_{(k+1):(k+h)}-M_h\left(X_{:k},\mathbf{y}_{:k}\right) \mathbf{w}(t_n)\Big)\\
        &=\arg\min_{\mathbf{w}(t_n)} \Bigg[ \frac{1}{2} \mathbf{w}(t_n)^T \left(\sum_{k = n-l_b}^{n-h}M_h(X_{:k},\mathbf{y}_{:k})^T M_h(X_{:k},\mathbf{y}_{:k})\right) \mathbf{w}(t_n)\\
        & \ \ \ \ -\left(\sum_{k = n-l_b}^{n-h} M_h(X_{:k},\mathbf{y}_{:k})^T \mathbf{y}_{(k+1):(k+h)} \right)^T \mathbf{w}(t_n) \Bigg] \\
        \\
        & \ \textrm{s.t.} \quad \mathbf{w}(t_n) \ge 0 \\
        & \ \ \ \ \ \ \ \  \mathbf{1}^T \cdot \mathbf{w}(t_n) = 1.
    \end{aligned}
\label{eq:optimization_problem}
\end{equation}

In fact, one can show that this is a so-called \textit{convex quadratic programming} optimization problem, for which efficient solvers exist \cite{nocedal2006numerical}. In particular, we apply the quadratic programming solver provided by the CVXOPT Python package \cite{cvxopt}. See Supplementary File for more details. Notice that the mean squared error is computed across the lock-back data set with stride equal to one ($s = 1$). The weights that solves \eqref{eq:optimization_problem} are denoted by $\hat{\mathbf{w}}(t_n)$. The predicted $h$-step-ahead forecasts at time $t_n$ is given by $M_h(X_{:n},\mathbf{y}_{:n}) \hat{\mathbf{w}}(t_n)$.

Since the ensemble is based on models trained on different time segments, and the weighting is based on the most recent dynamics, we call the method the \textit{Recency-Weighted Temporally-Segmented} (ReWTS) ensemble method.
As opposed to in \cite{wang_mining_2003}, the weights are not computed separately for each chunk model. Instead, by the construction of the optimization problem in \ref{eq:optimization_problem}, the best performing chunk models on the look-back data will be assigned the largest weights implicitly due to the constraint that the weights need to sum to one. The ReWTS ensemble is arguably more similar to the ensemble model described in \cite{galicia_multi-step_2019}. However, in \cite{galicia_multi-step_2019}, all the models are trained on the same data, specifically restricted to the most recent data. Hence, in the context of data streams, previous relationships are disregarded, and only the most recent data is considered relevant. This approach risks forgetting or never learning about recurring events and cyclic patterns, which are often observed in industrial processes. 

Like the Mixture of Experts (MoE) model \cite{masoudnia2014mixture}, our model facilitates model specialization and employs a weighting strategy during inference to determine the most relevant subset of models for a given input. However, in our approach, specialization is achieved through explicit segmentation of training data, whereas MoE employs a learned gating network to direct inputs to specific experts. At inference, MoE uses this gating network to activate relevant experts based on current input, whereas ReWTS employs a structured optimization process informed by recent historical data to select operative models. Unlike MoE, which primarily aims to enhance computational efficiency, ReWTS is designed to improve forecasting accuracy in complex time-series data.

\subsection{The entire forecasting process}\label{forecasting}
Consider a defined model architecture with fixed hyperparameters for each chunk model, chunk length $l_c$, look-back length $l_b$, and forecast horizon $h$. Let $t_n$ be the present time point, and assume that we have just finished chunk $c$ (i.e.\ $n/l_c$ is integer) with $c$ trained models $\mathcal{M}_{n} = \{M_1,\ldots,M_c\}$. For ease of notation, and to avoid any further complexity, we will assume that the time it requires to train each model is less than the time interval between two consecutive time points, $\Delta t$. With a continuous flow of new data, the automatic forecasting process to be employed in practice can be described as given in Algorithm \ref{alg:chunk_model_forecasting}:

\begin{algorithm}
\caption{The $h$ step forecasting procedure - ReWTS ensemble model}
\begin{algorithmic}[1]
    \STATE $\triangleright$ Given forecast horizon $h$, look-back length $l_b$, chunk length 
 $l_c$, stride $s$ and present time point $t_n$, with $n/l_c$ an integer
    \STATE $\triangleright$ Assign $C = n/l_c$, the number of available chunks
    \STATE $\triangleright$ Assign $v = n$
    \STATE $\triangleright$ Given all previously trained models, $\mathcal{M}_{t_v} = \{M_1,...,M_C\}$, from the $C$ previous chunks
    
    \REPEAT
        \WHILE{$v \le (C+1) \cdot l_c$}
        \STATE $\triangleright$ Collect look-back data $(X_{(v-l_b):v},\mathbf{y}_{(v-l_b):v})$ 
        \STATE $\triangleright$ Calculate the weights $\hat{\mathbf{w}}(t_v)$ according to optimization problem \eqref{eq:optimization_problem} given chunk models in  $\mathcal{M}_{v}$
        \STATE $\triangleright$ Compute $h$-step forecasts from time $t_v$ for all $c$ chunk models, and save in matrix $M_h(X_{:v},\mathbf{y}_{:v})$
        \STATE $\triangleright$ Compute final forecasts given by $M_h(X_{:v},\mathbf{y}_{:v}) \hat{\mathbf{w}}(t_v)$ at present time point $t_v$
        \STATE $\triangleright$ $v = v + s$
        \ENDWHILE
        \STATE $\triangleright$ Collect all data from new chunk number $C+1$ between time points $t_{C \cdot l_c}$ and $t_{(C+1) \cdot l_c}$
        \STATE $\triangleright$ Train new chunk model $M_{C+1}$ from chunk $C+1$ and update $\mathcal{M}_{v}$
        \STATE $\triangleright$ $C = C + 1$ 
    \UNTIL{Collection of new data stops}
\end{algorithmic}
\label{alg:chunk_model_forecasting}
\end{algorithm}

It is important to have in mind that the data generating process can change over time, and thus be different from chunk to chunk. Before any chunk model is fitted, feature scaling such as standardization is often applied. Before computing forecasts using a previously fitted chunk model on new data with an unknown data generating process, it is hence crucial to first scale the input data to the model according to the observed data generating process in the previously seen chunk. 
     
\begin{figure}[h]
\centering
\includegraphics[width=16cm]{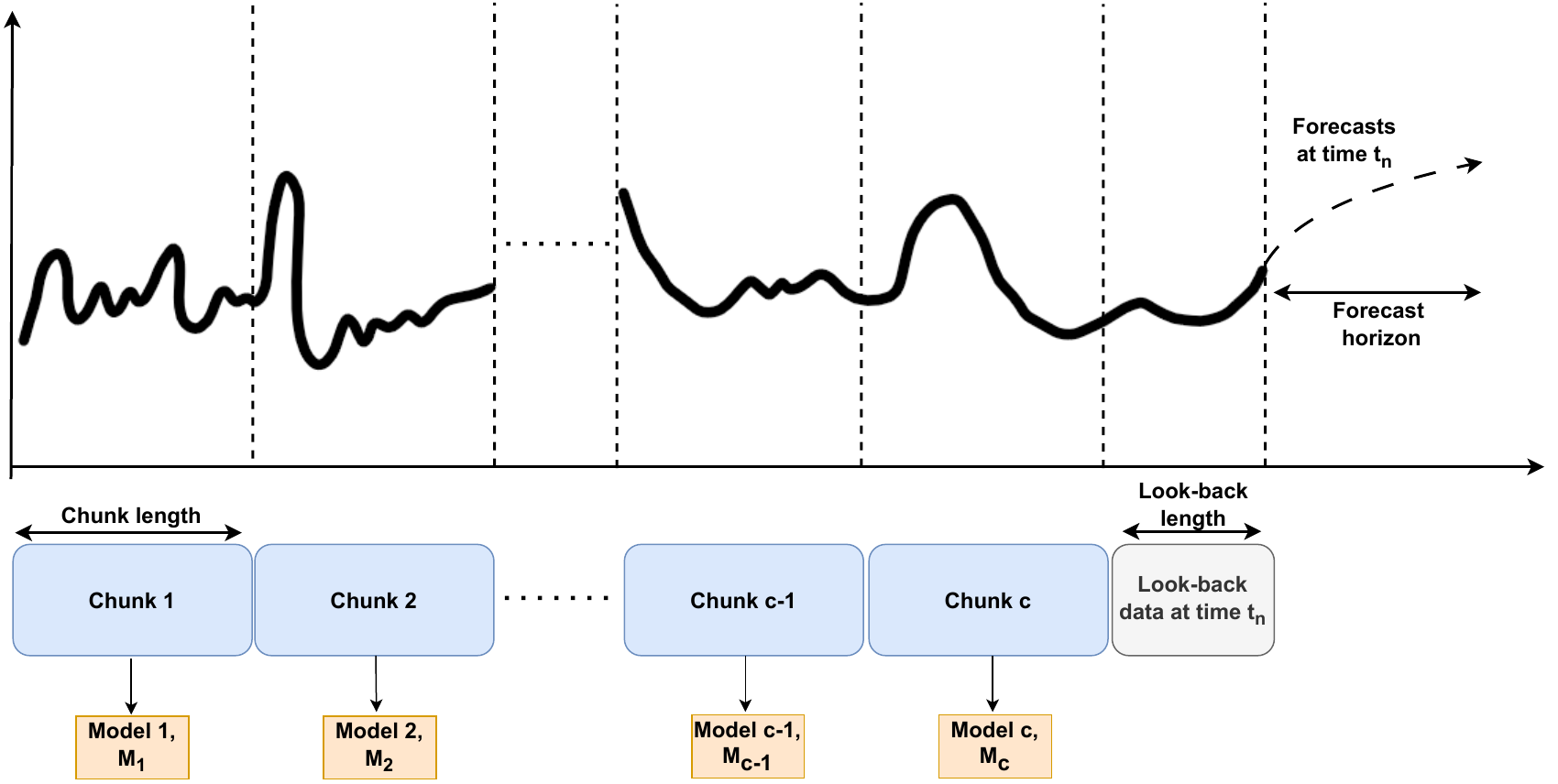}
\caption{Conceptual illustration of the look-back, chunk-based approach. The figure shows how the models trained on disjoint chunks of times series are used to provide forecasts of the target response at time $t$ given look-back data.}
\label{fig:concept_fig}
\end{figure}

Figure \ref{fig:concept_fig} provides a schematic overview of the proposed ReWTS ensemble model approach using a look-back data set.

\section{Methodology}\label{sec:methodology}

\subsection{Architectures and hyperparameters}
The implementation of the ReWTS ensemble model is integrated within the Python package Darts \cite{darts}. In this way, one can easily experiment with several built-in model architectures.
In this work, each model in the ReWTS ensemble model will be of the same machine learning architecture, and trained on separate disjoint time intervals of equal size as depicted in Figure \ref{fig:concept_fig}. However, we test various architectures including both neural networks and tree-based models. 

There are several choices to be made regarding hyperparameters that must be in place before deploying the automatic forecasting process of the ReWTS ensemble model. This includes model-specific hyperparameters (such as learning rate, hidden layers etc.) as well as chunk length and look-back length. Our software package leverages the Hydra framework \cite{Yadan2019Hydra}, enhancing modularity and flexibility to streamline configuration management and enable rapid experimentation. It further features a pipeline using the Optuna package for hyperparameter optimization. \cite{optuna}. The source code for this work is openly available at \url{https://github.com/SINTEF/rewts}.

To investigate the ReWTS ensemble model, we employ it on generated simulation data, as well as based on real process industry data. 

\subsection{Baseline model - The global model}\label{sec:full_model_forecasting}
For both the simulated data and the process industry data, we compare the ReWTS ensemble model with what we call a \textit{global model}, which means that one single forecasting model is trained on all historic data at once.

The automatic forecasting process for the global model is described in Algorithm \ref{alg:full_model_forecasting} given a model architecture and predefined hyperparameters. This means that the global model is retrained from scratch for every new chunk of data. The smaller chunk length, the more often the global model is updated.

\begin{algorithm}
\caption{The $h$ step forecasting procedure - Global model}
\begin{algorithmic}[1]
    \STATE $\triangleright$ Given forecast horizon, $h$, chunk length, 
 $l_c$, stride $s$ and present time point $t_n$ with $n/l_c$ an integer
    \STATE $\triangleright$ Given previously trained global model, $M_f$ using all historic data up to time $t_n$
    \STATE $\triangleright$ Assign $v = n$
    \STATE $\triangleright$ Assign $C = n/l_c$, the number of available chunks
    \REPEAT
        \WHILE{$v \le (C+1) \cdot l_c$}
       
        \STATE $\triangleright$ Compute $h$ step forecasts from global model given by $M_f(X_{:v},\mathbf{y}_{:v})$ at present time point $t_v$
        \STATE $\triangleright$ $v = v + s$
        \ENDWHILE
        \STATE $\triangleright$ Collect all data from new chunk number $C+1$ between time points $t_{C \cdot l_c}$ and $t_{(C+1) \cdot l_c}$
        \STATE $\triangleright$ Update global model $M_{f}$ from scratch using all historic data including new data from chunk $C+1$
        \STATE $\triangleright$ $C = C + 1$ 
    \UNTIL{Collection of new data stops}
\end{algorithmic}
\label{alg:full_model_forecasting}
\end{algorithm}

The baseline model has the same model architecture as the individual chunk models in the ReWTS ensemble. However, in the pursuit of fair evaluation, the ReWTS ensemble is compared to a slightly altered version of the global model for neural network models. In this case, we make sure that during forecasting, the ReWTS ensemble model and the global model, at every time point, has the same number of trainable parameters. In practice this means that prior to retraining of the global model, the dimensions of the layers in the neural network are upscaled according to the dimensions of the new chunk model added to the ensemble model. This is not done for other model architectures, such as linear models or tree ensemble models. For such models, ensuring the same number of trainable parameters for the global model and ReWTS ensemble model is not necessarily beneficial for the corresponding global model, due to potential overfitting. Therefore, the number of trainable parameters are the same as in each chunk model.

For both version of the global model, initial HPO before deployment of the automatic forecasting process is based solely on the global model. The resulting fitted hyperparameter values from the global model are used directly to assign hyperparameter values for each chunk model. More details on this follow in Section \ref{sec:sim_data_description} and \ref{process_data}.

\subsection{Simulation data}\label{sec:sim_data_description}
To demonstrate the potential advantages of the ReWTS ensemble model and identify the conditions where it proves beneficial, we create a synthetic dataset. The synthetic dataset has distinct data-generating processes, in terms of sinusoidal functions, in non-overlapping consecutive regions or chunks. Hence, each chunk has its unique data pattern, creating well-defined transitions between them. This design choice allows us to assign one model per chunk, ensuring each model in the ReWTS ensemble is trained exclusively on data from one data generating process.

The synthetic dataset is made up of 16 chunks each consisting of 500 data points sampled from a sine wave ($A \sin (\omega t)$), where the amplitudes and frequencies are given in Table \ref{tab:sine:dataset_params}, while making sure the time series is continuous between all chunks. We denote the first eight chunks to be in the train dataset, and the rest in the test dataset.
\begin{table}
    \centering
    \begin{tabular}{lccccccccc}
        \toprule
        \multirow{2}{*}{Dataset} & \multirow{2}{*}{Parameter} & \multicolumn{8}{c}{Chunk} \\
        \cmidrule{3-10} & & \#1 & \#2 & \#3 & \#4 & \#5 & \#6 & \#7 & \#8 \\
        \midrule
        \multirow{2}{*}{Train} & Amplitude $A$ & 0.5 & 2 & 20 & 2 & 2 & 0.5 & 2 & 5 \\
        & Frequency $\omega$ & 10 & 2 & 5 & 1 & 0.5 & 8 & 3 & 1 \\
        \midrule
        \multirow{2}{*}{Test} & Amplitude $A$ & 0.75 & 10 & 3 & 0.5 & 5 & 1.25 & 3 & 4 \\
        & Frequency $\omega$ & 8 & 0.75 & 7 & 11 & 0.65 & 4 & 2 & 5 \\
        \bottomrule
    \end{tabular}
    \caption{Parameters used to generate the various chunks of the synthetic dataset.}
    \label{tab:sine:dataset_params}
\end{table}

Our hypothesis posits that the ReWTS ensemble model in these circumstances should perform better in total than the corresponding global model. This is because each chunk model in the ReWTSensemble model can specialize in one specific data generating process, while the global model will focus its effort on learning the chunks with dominating dynamics (i.e. in this case the largest amplitudes). This is advantageous in these specific chunks, yet it yields less focus to generalize for all chunks. 

As an interesting feature: When forecasting with the ReWTS ensemble model, the weight assignment procedure should ideally allocate all weight to the model trained on the same data generating process for which we are currently forecasting, assuming the occurrence of the data generating process in question has previously taken place.

For each chunk in the train dataset we train one forecasting model until convergence on a hold-out validation set accounting for the last $25\%$ of the chunk. We then construct a ReWTS ensemble model of the resulting eight models. The look-back length is set to $l_b = 160$ time steps. The forecasts of the ReWTS ensemble model at any time point will then be a linear combination of these eight models, where the corresponding weights are calculated as given in Algorithm \ref{alg:chunk_model_forecasting}. The ReWTS ensemble's performance on the train and test datasets is compared against the global model that is trained simultanously on all eight chunks in the training set. For both the ReWTS ensemble model and the global model, a vanilla recurrent neural network (RNN), specifically of type Long-short-term memory (LSTM) \cite{staudemeyer_understanding_2019}, model architecture is applied. Both the LSTM model for each chunk model, and the global LSTM model has an input length of 80 time steps, and otherwise equal hyperparameter values such as optimizer (Adam), batch size, early stopping and max epochs. The difference lies in the dimension space of the hidden LSTM layer. As first mentioned in Section \ref{sec:full_model_forecasting}, to compare the ReWTS ensemble with a global model with comparable learning capacities, we make sure the global model has as many learnable parameters as the total number of parameters in the ReWTS ensemble model. Therefore, each chunk model is set to have a single LSTM layer of dimension 10, yielding $\approx 500$ learnable parameters. With eight chunk models, this implies the global model to have around eight times more parameters than each chunk model, 4000 in total, resulting in an LSTM layer of dimension 32.

We deploy the automatic forecasting process for the chunk model and the global model as described in respectively Section \ref{sec:Chunk} and \ref{sec:methodology} on the train data as well as the test data with unseen sinusoidal functions. The forecasts are evaluated based on their ability to forecast 30 time steps ahead for every 30th time step (stride of 30). We evaluate the performance of both methods in each chunk separately using the loss function given in \eqref{loss}, for all forecasts between $[t_{i \cdot l_c},t_{(i+1) \cdot l_c}]$, with $t_{i \cdot l_c}$ denoting first time step of chunk $i$. To make the performances on the different chunks comparable, we further scale by the amplitude of the corresponding sine wave in the chunk.
Separately from the forecasting evaluations explained above, we will also investigate the behaviour of the two approaches in the transitions between the chunks where there is a sudden shift in the data generating process.

\subsection{Process industry data}\label{process_data}
We investigate the performance of the ReWTS ensemble model on near two-year long real process industry data from two water treatment plants in Norway. Specifically, we utilize one dataset spanning from January 2021 to January 2023 gathered from a drinking water treatment plant managed by Bergen Vann, the city of Bergen's drinking water producer. Additionally, we utilize another dataset collected between January 2022 and October 2023 from Veas, a wastewater treatment plant in Slemmestad, responsible for processing wastewater originating from the Oslo region.

\subsubsection{Wastewater treatment data}\label{sec:veas}
\begin{figure}[h]
\centering
\includegraphics{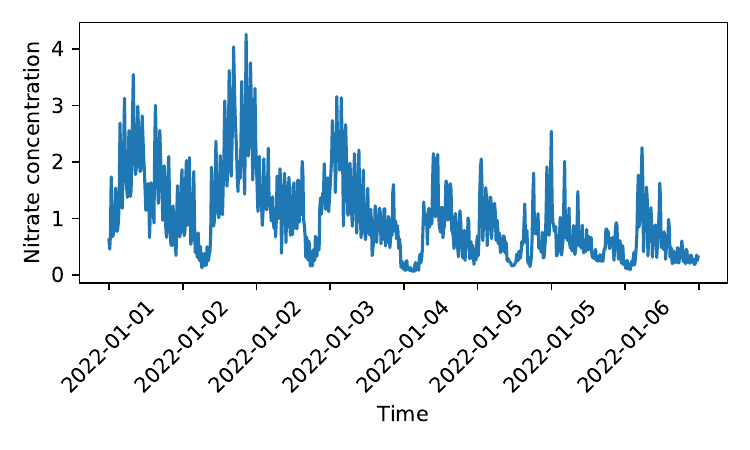}
\caption{A small time segment of the first six days in January 2022 showing how the nitrate concentration in the wastewater plant varies in time. Time resolution is in 10 minutes.}
\label{fig:veas}
\end{figure}

For the wastewater treatment plant, we investigate the process of denitrification where facultative anaerobic bacteria naturally present in the wastewater convert nitrate into dinitrogen. These bacteria respire nitrate in low-oxygen conditions. The conversion is achieved using a carbon and electron source, which for the particular wastewater treatment plan is methanol. Removing nitrogen is one of the main objectives of the wastewater treatment plant, and accurate modeling of the nitrate concentration out of the process is important for optimal dosage of methanol. We focus on a single hall consisting of four filters in parallel. The historic data is resampled to have a frequency of 10 minutes between every measurement, which will also be the basis when developing the forecasting models. Figure \ref{fig:veas} shows a visualization of a typical trend of the nitrate concentration after passing the denitrification hall as a function of time. One can observe a large non-stationary fluctuation of the nitrate concentration. The measurements are also noisy in nature as the measured nitrate concentration is based on samples from four separate filters, and these filters can be in different states. The use case for this data set is to provide forecasts for the nitrate concentration after the denitrification at intervals of 10 minutes, spanning up to four hours into the future. Sensor measurements given as input to the model is the nitrate concentration before entering the denitrification hall, the flow of methanol, as well as the measured nitrate concentration in the wastewater after being process in the denitrification hall. Before constructing the forecasting models, measurements of the nitrate concentration out of the hall during a small time interval were removed due to severe measurement errors. These measurements were replaced by linearly interpolated values. 

\subsubsection{Drinking water treatment data}\label{sec:bergen}
For the drinking water treatment plant, we investigate one particular filter, out of several, that cleans drinking water. The data consists of many different measurements. This includes the turbidity of the drinking water, a measure of how cloudy the water is. Additionally, we have drinking water produced per time unit, amount of raw water entering the treatment plant per time unit, pH, temperature in water, conductivity, chemical dosing of iron chloride as well as measurements connected to regular flushing of the filter, among them flushing water amount, backwash air, maturation rate and maturation time. The raw data, which is sampled at one minute time resolution, is resampled to a frequency of 10 minutes. The target variable to forecast is the turbidity in the drinking water measured in Formazin Nephelometric Units. The aim is to provide forecasts at intervals of 10 minutes spanning up to four hours into the future. Between mid-September and mid-November in 2022 there is a period with error measurements in the turbidity due to planned maintenance, in particular renewal of filtration media. This is a typical and inevitable scenario in process industries, and it is further difficult for the forecasting models to avoid using erroneous measurements as training data at all times when operating online. We therefore leave the incorrect measurements in the data to explore how the global model and the ReWTS ensemble model behave in these circumstances.

\begin{figure}[h]
\centering
\includegraphics{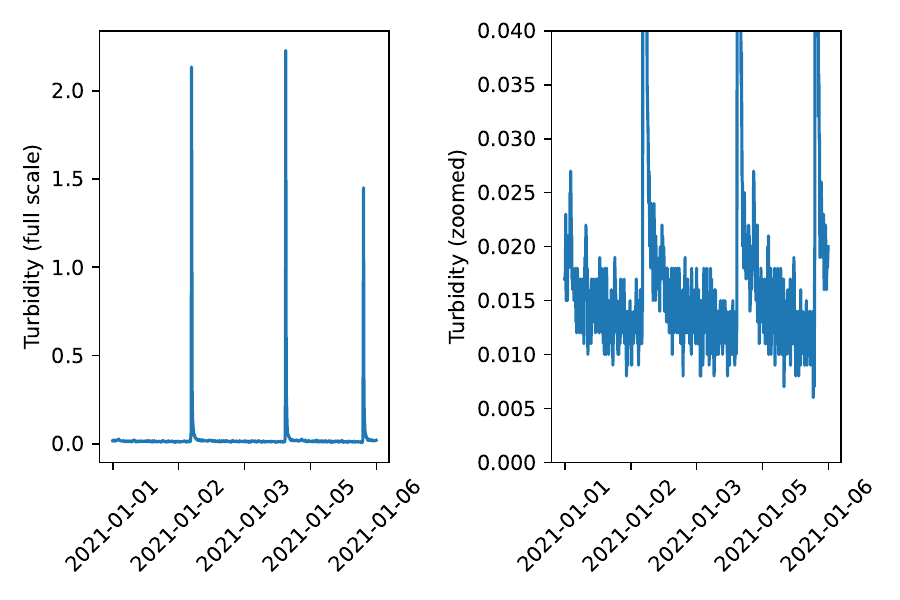}
\caption{Figure to the left: A small time segment of the first six days in January 2021 showing how the turbidity at the drinking water plant varies in time. The peaks in turbidity indicates when the filter is flushed. Drinking water is produced between the peaks. Time resolution is 10 minutes. Figure to the right: Measurements for the same time segment, but zoomed in to see the dynamics of the turbidity during drinking water production..}
\label{fig:bergen}
\end{figure}

A visualization of a typical trend of the turbidity as a function of time is given in Figure \ref{fig:bergen}, where one can observe regular spikes in the turbidity measurements, which corresponds to time intervals where the filter is cleaned. In the time intervals between the spikes, drinking water is produced at required turbidity levels. 

\subsubsection{Online prediction framework and initial hyperparameter selection}\label{hyp_opt_process_data}

Unlike in the simulation data, we typically do not know exactly the time spans of different data generating processes in real process industry data. Hence, in addition to model-specific hyperparameters, chunk length and look-back length needs to be determined before deployment of an automatic forecasting process. In this work we evaluate these hyperparameters by exploring the effect on forecast performance for different values of one parameter while holding the other fixed and vice versa (see Figures \ref{fig:veas_results_vary_chunk_length},\ref{fig:veas_results_vary_lookback_length}). To decide model-specific hyperparameters we reserve the first third (33 \%) of the data (around half a year) for HPO, using the first 90\% to train models and the last 10\% to evaluate the hyperparameter choices. In pursuit of building a strong baseline, during HPO we train and evaluate hyperparameters in the global model fashion such that the baseline model we compare against should be as optimal as possible. We then apply the same hyperparameters resulting from HPO of the global model as input for each chunk model in the ensemble, with the exception of hyperparameters deciding the number of tunable parameters. In this case, we scale down such that the corresponding ensemble generated from the HPO data as a whole has equal number of tunable parameters to the global model, given the specified chunk length. More details regarding this are given in Section \ref{sec:forecasting_process} below. The hyperparameters resulting from this HPO procedure are specified in the files configs/model/\{veas,bergen\}\_\{rnn,tcn,xgboost,elastic\_net\}.yaml in the project code-base available at \url{https://github.com/SINTEF/rewts}.

We investigate four different model architectures: a linear model with elastic net regularization \cite{zou_regularization_2005}, the tree ensemble boosting model XGBoost \cite{chen_xgboost_2016}, a temporal convolutional network (TCN) \cite{bai_empirical_2018} and a LSTM model \cite{staudemeyer_understanding_2019}. All model architectures are directly available through the Darts python package. These models have in common that the corresponding input data to the model includes both historic values of the target variable as well as other covariates that are thought to have an effect on the target variable. The covariates used in the model are assumed known in the future when computing the forecasts. As mentioned earlier, this is often the case in process industries where control signals can be fully determined such as future dosing of chemicals. All the chosen model architectures can be applied \textit{autoregressively} to produce multi-step forecasts, however, note that the HPO procedure described above resulted in the linear, XGBoost, and LSTM model all outputting one-step predictions at a time, while the TCN model outputs several steps at once. 

\subsubsection{Forecasting on new incoming data}\label{sec:forecasting_process}
In this section we outline the automatic forecasting procedure for the global and ReWTS ensemble model for a particular process industry dataset with measurements in the interval $[0,t_e]$. Given a model architecture, model-specific HPO of the global model is executed as explained in Section \ref{hyp_opt_process_data} using the data in the interval $[0,t_e/3]$. Thereafter, chunk length, $l_c$, and look-back length, $l_b$, for the ReWTS ensemble model is set, which gives $\lambda= \lfloor t_e/3l_c \rfloor$ chunk models within the interval $[0,t_e/3]$. If the model architecture is a neural network (TCN or LSTM), let $p$ denote the fitted number of trainable parameters as a result of the chosen hyperparameters from HPO. To facilitate a fair comparison, we ensure that the global model at all times has at least as many trainable parameters as the sum of trainable parameters of the ReWTS ensemble model. This is achieved by downscaling the hyperparameters affecting the number of trainable parameters of each chunk model such that the number of trainable parameters are approximately $p/\lambda$. For model architectures of type linear and XGBoost, the hyperparameter values of each chunk model will be equal to those hyperparameter values found during HPO of the global model. At this point, all hyperparameter values are set.

The automatic forecasting process for the ReWTS ensemble model is carried out as explained in Section \ref{forecasting} with $C = 2$. That is, we start the forecasting when we initially have trained two chunk models from the chunks $[0,t_{l_c}]$ and $[t_{l_c},t_{2 l_c}]$ respectively, with $p/\lambda$ trainable parameters in each chunk.

The automatic forecasting process for the global model, with $C = 2$, is carried out as described in \ref{sec:full_model_forecasting}. Hence, the initial global model is trained on data from both chunks on the whole time interval $[0,t_{2 l_c}]$. The global model is the same type of model as for each chunk model. For the linear model and XGBoost model architecture, the hyperparameter values are identical. For the LSTM and TCN model architectures, the number of trainable parameters of the global model is equal to $2 p/\lambda$, such that it matches the total number of trainable parameters of the corresponding ReWTS ensemble. The global model is retrained from scratch from every new chunk of data of length $l_c$ using all historic data. When retraining the TCN and LSTM model, the number of trainable parameters is adaptively upscaled from $(C-1) p/\lambda$ to $C p/\lambda$ when finishing chunk number $C$.

For comparison, we will also compare the ReWTS ensemble with a static global model without upscaling of trainable weights. Specifically, the constructed global model after HPO with $p$ trainable parameters is directly applied in Algorithm \ref{alg:full_model_forecasting} with $C=2$. Hence, the retraining includes the data from the most recent chunk, but the number of trainable parameters, $p$, is kept constant.

We compare the performance of the automatic forecasting process for the global model and the ReWTS ensemble model respectively for each chunk using the loss function given in Equation \eqref{loss}. The total performance over the whole dataset is measured as the mean performance over all chunks. Finally, to see the effect of the choice of chunk length and look-back length, we repeat the process above for different values of chunk length, and look-back length. In particular, we investigate the chunk lengths 5, 7, 14, 21 and 28 days while keeping the look-back length fixed at 3 days. The look-back length is investigated for 2, 3, 5, 7 and 10 days while keeping the chunk length at 14 days.

\section{Results} \label{sec:Results}

\subsection{Simulation data}

\begin{figure}[t!]
    \centering
    \begin{subfigure}{\textwidth}
        \centering
        \includegraphics{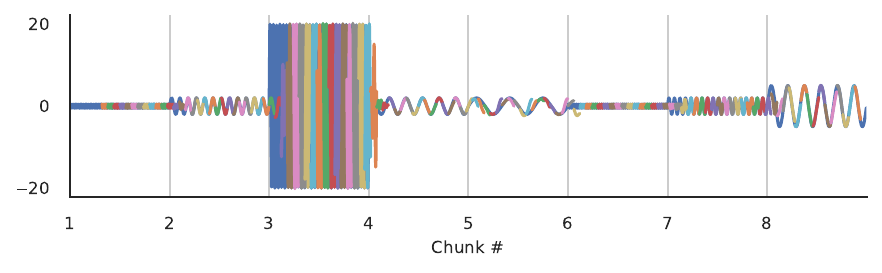}
        \caption{Ensemble model: Training set}
        \label{fig:synthetic_chunk_train}
    \end{subfigure}
    
    \begin{subfigure}{\textwidth}
        \centering
        \includegraphics{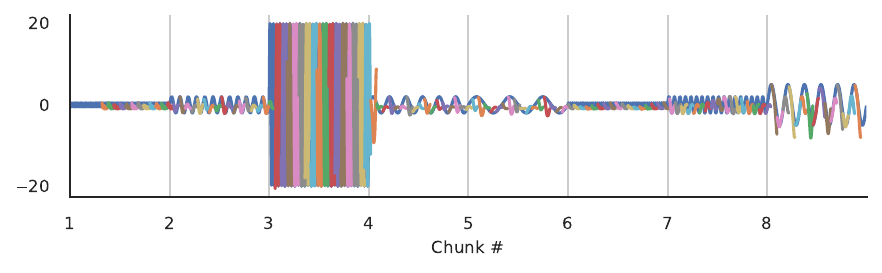}
        \caption{Global model: Training set}
        \label{fig:synthetic_full_train}
    \end{subfigure}

    \begin{subfigure}{\textwidth}
        \centering
        \includegraphics{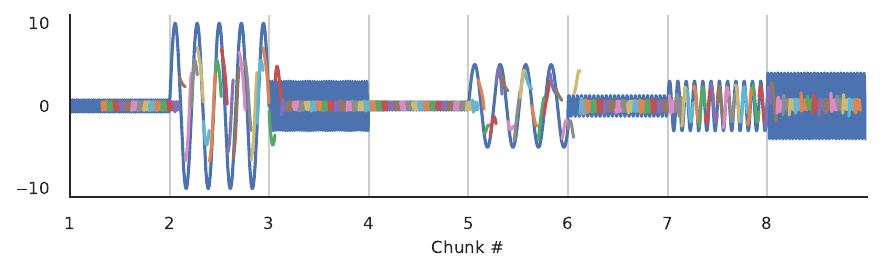}
        \caption{Ensemble model: Test set}
        \label{fig:synthetic_chunk_test}
    \end{subfigure}
    
    \begin{subfigure}{\textwidth}
        \centering
        \includegraphics{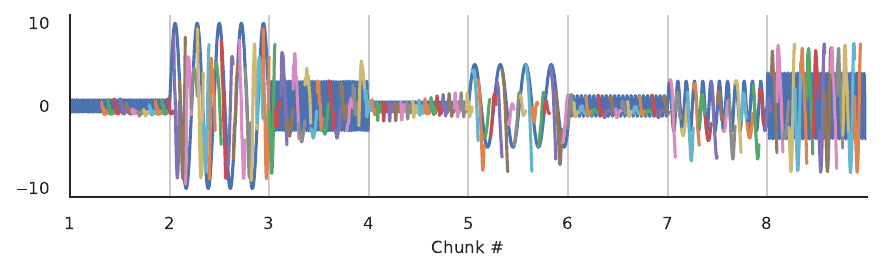}
        \caption{Global model: Test set}
        \label{fig:synthetic_full_test}
    \end{subfigure}
    \caption{Data (blue) and model predictions (colors) for all chunks for the training set and the test data from the sine experiment. Note how the global model has prioritized learning the dynamics of the dominant amplitude in chunk \#3 of the training set.}
    \label{fig:synthetic_all}
\end{figure}

We evaluate our proposed ReWTS ensemble model by using the simulated data as described in Section \ref{sec:sim_data_description}. A summary of the results are given in Table \ref{tab:sine:results} and Figure \ref{fig:synthetic_all} showing the performance of both the ReWTS ensemble model and the global model according to the loss function given in \ref{loss} for each chunk individually as well for all chunks as a whole. The ReWTS ensemble model generally outperforms the global model in the train dataset as well as on the unseen test dataset.

\begin{table}[h]
    \centering
    \resizebox{\columnwidth}{!}{%
    \begin{tabular}{llccccccccc}
        \toprule
        \multirow{2}{*}{Dataset} & \multirow{2}{*}{Model} & \multicolumn{9}{c}{Chunk} \\
        \cmidrule(lr){3-11}
        & & \#1 & \#2 & \#3 & \#4 & \#5 & \#6 & \#7 & \#8 & Mean \\
        \midrule
        \multirow{2}{*}{\centering Train} & ReWTS & {\bf 1.97E-05} & {\bf 6.04E-04} & 4.86E-03 & {\bf 9.41E-03} & {\bf 2.00E-02} & {\bf 4.15E-05} & {\bf 3.95E-04} & {\bf 2.37E-02} & {\bf 7.38E-03} \\
        & Global                          & 7.49E-01 & 6.13E-01 & {\bf 1.47E-03} & 1.18E+00 & 1.38E+00 & 1.33E+00 & 6.90E-01 & 1.20E+00 & 8.94E-01 \\
        \midrule
        \multirow{2}{*}{\centering Test} & ReWTS & {\bf 5.90E-02} & {\bf 1.74E+00} & {\bf 1.20E+00} & {\bf 2.46E-01} & {\bf 7.57E-01} & {\bf 4.87E-01} & {\bf 2.02E-01} & {\bf 8.36E-01} & {\bf 6.90E-01} \\
        & Global & 1.08E+00 & 6.15E+00 & 4.55E+00 & 1.11E+00 & 2.05E+00 & 9.47E-01 & 1.87E+00 & 7.28E+00 & 3.13E+00 \\
        \bottomrule
    \end{tabular}%
    }
    \caption{Relative (normalized by the maximum amplitude value) mean squared error from applying the ReWTS and global approaches on the simulation data. }
    \label{tab:sine:results}
\end{table}

This holds true when evaluating over all concatenated chunks as one time series (Figure \ref{fig:synthetic_all}), and when assessing each chunk individually, except for Chunk \#3 in the training dataset where the global model attains a smaller loss. See Supplementary File to get a closer look at the predicted forecasts from the ReWTS ensemble model and the global model for each of the 16 chunks. This behaviour follows our hypothesis, where the global model does exceptionally well on the chunk with the highest amplitude that dominates the learning objective, at the cost of performance on the other data chunks. Furthermore, the ReWTS ensemble model demonstrates superior extrapolation abilities compared to the global model with respect to the results on the test data.

\begin{figure}[h!]
    \centering
    \begin{subfigure}{0.49\textwidth}
        \centering
        \includegraphics{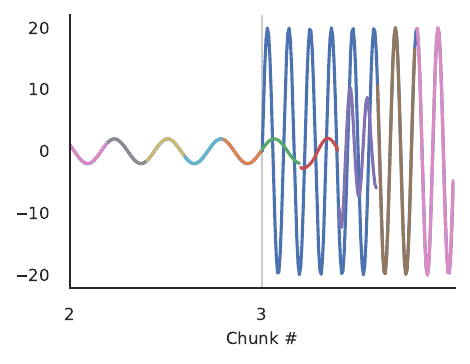}
        \caption{Ensemble model}
        \label{fig:chunk_edge}
    \end{subfigure}
    \hfill
    \begin{subfigure}{0.49\textwidth}
        \centering
        \includegraphics{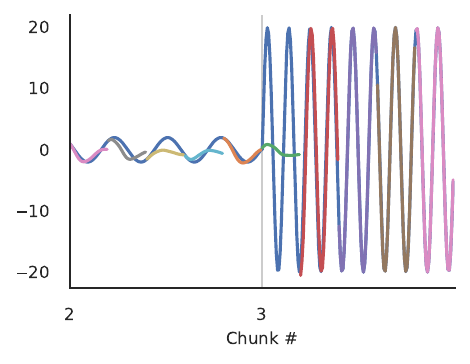}
        \caption{Global model}
        \label{fig:full_edge}
    \end{subfigure}
    \caption{With rapid change in the dynamics, the ReWTS ensemble model is slow to adapt (the true dynamics given by the blue line).}
    \label{fig:synthetic:edge}
\end{figure}

Note that when evaluating the ReWTS ensemble model over concatenated chunks, its adjustment to a new data generating process is more gradual compared to the global model. This behavior is due to the weight assignment procedure based on the look-back data preceding the current prediction point. As a result, the look-back data will include information from the previous chunk, slowing the ReWTS ensemble's adjustment, until we are further than the length of the look-back data into the new chunk of data generating process. These edge effects are highlighted in Figure \ref{fig:synthetic:edge}, where the global model can be seen to be better at adapting to a sharp change in the dynamics.

The evaluation on the simulation data has been based on an autoregressive LSTM model. As explained in Section \ref{sec:Chunk}, the weights are calculated based on the performance of the $h$-step forecasts for each model on the look-back data. This is intuitive to do since the final task is to predict $h$-step forecasts at the present time point, and hence the evaluation of the chunk models in the look-back data should reflect what is the final task. An alternative procedure for autoregressive models is to find the weights that minimizes the \textit{one-step} ahead forecast in the look-back data, equivalent to setting $h =1$ in the QP optimization problem \eqref{eq:optimization_problem}. At prediction time, the $h$-step forecasts ($h>1$) can in this case be computed by reapplying the computed weights $h$ times recursively. See Supplementary file for more algorithmic details. More generally speaking, the forecast horizon applied in Algorithm \ref{alg:chunk_model_forecasting} does not need to be the same as the desired forecasting horizon, $h$, at prediction time. In fact, the forecast horizon used in Algorithm \ref{alg:chunk_model_forecasting} can be smaller than $h$, and the corresponding weights can be reapplied recursively the number of times necessary to provide $h$-step forecasts.

\begin{figure}[t!]
    \centering
    \begin{subfigure}{0.45\textwidth}
        \centering
        \includegraphics[width=\linewidth]{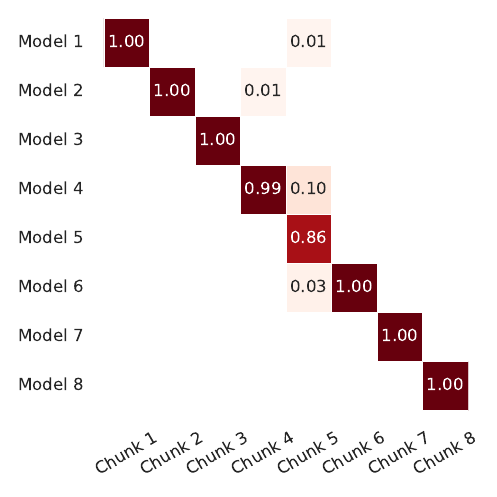}
        \caption{Weights $h=30$}
        \label{fig:h-step}
    \end{subfigure}
    \begin{subfigure}{0.45\textwidth}
        \centering
        \includegraphics[width=\linewidth]{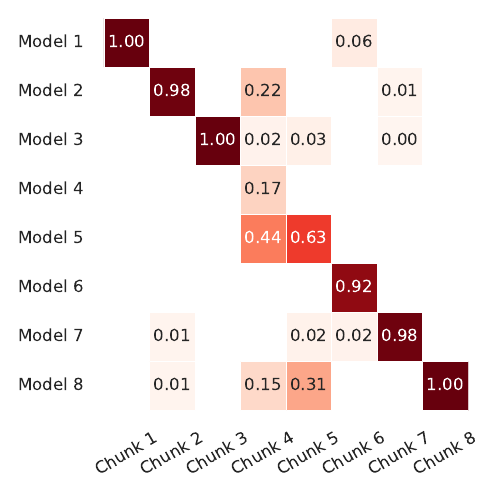}
        \caption{Weights $h=1$}
        \label{fig:one-step}
    \end{subfigure}
    \caption{Increasing the forecast horizon when fitting ensemble weights increases accuracy with which it identifies the most suitable model.}
    \label{fig:h-step_vs_one-step}
\end{figure}

For the simulation data, and for the evaluation of the ReWTS ensemble model on the training data, we ideally want the ReWTS ensemble model to recognize the previously seen pattern at prediction time, corresponding to the dynamics in the present chunk. This means that we want the model that has been trained on a particular chunk to be assigned a weight equal to one at prediction time, should the exact same dynamics in the chunk reappear in the future. In Figure \ref{fig:h-step_vs_one-step}, we investigate the distribution of the weights for each model for each chunk, for two different weight assignments. The first weight assignment is where the forecast horizon in Algorithm \ref{alg:chunk_model_forecasting} is equal to the forecast horizon at prediction time, $h=30$. The second weight assignment is where the forecast horizon in Algorithm \ref{alg:chunk_model_forecasting} is equal to one, and where the weights are reapplied $h$ times at prediction time. We see that the first weight assignment ($h=30$, Figure \ref{fig:h-step}), outperforms the second weight assignment ($h=1$, Figure \ref{fig:one-step}), in terms of allocating the correct chunk model with the largest weight.

\begin{figure}[b!]
    \centering
    \includegraphics{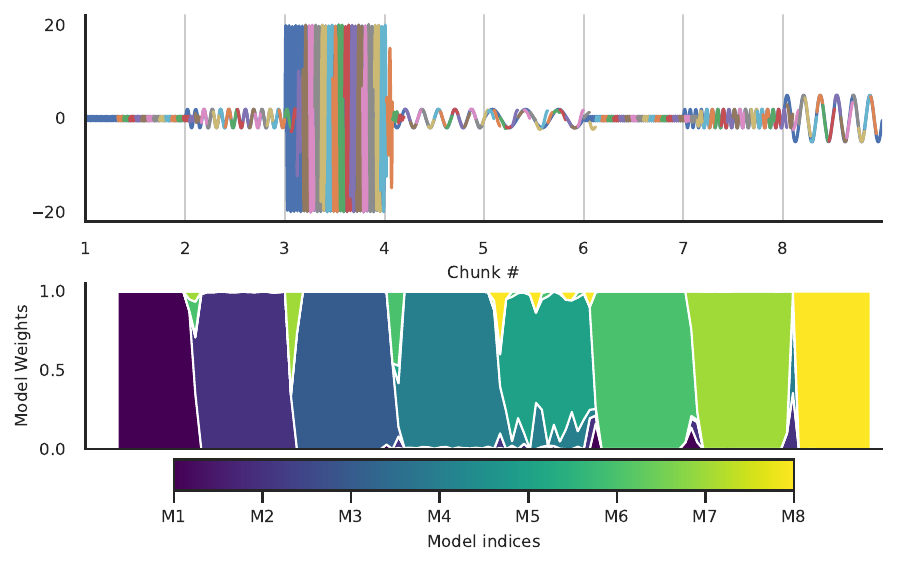}

    \caption{Ensemble model predictions on the train set along with the weights assigned to the models.}
    \label{fig:synthetic:ensemble_weights}
\end{figure}

By the construction of the ReWTS ensemble model, the purpose of the weight assignments of the chunk models goes beyond the goal of achieving better forecasts than a corresponding global model. In fact, the weights can be used for interpretation, such as identifying previous intervals that are considered relevant at prediction time. Figure \ref{fig:synthetic:ensemble_weights} shows how the weights are distributed at different prediction times for the training set, when applying Algorithm \ref{alg:chunk_model_forecasting} with $h=30$. Inside the chunks, the correct chunk model is assigned a large weight close to or equal to one. To evaluate the trustworthiness of a particular forecast, one way to proceed is to investigate the properties of the data distributions of the chunk data sets corresponding to the chunk models with the largest weights. This can be a valuable property in the context where detection of recurring events is relevant.

In Figure \ref{fig:synthetic:ensemble_weights} we also observe the previously discussed edge effects between two neighbouring chunks. Another interesting observation is at chunk 5 where the ReWTS ensemble model is struggling the most to have a weight distribution concentrated around the correct chunk model. By closer inspection, one can see that the chunk model for chunk 4 is assigned a weight of 0.1. This is reasonable given the close resemblance between the dynamics in chunk 4 and 5.

\subsection{Process industry data}
 We present the evaluation of the ReWTS ensemble model based on the real process industry data from both Veas wastewater treatment plant and the drinking water treatment plant in Bergen. Based on the results from the simulation data, the ReTWS ensemble models are trained using Algorithm \ref{alg:chunk_model_forecasting} with the parameter $h$ equal to the desired forecasting horizon at prediction time.

\subsubsection{Results on wastewater treatment plant}

\begin{figure}[h!]
    \centering
    \begin{subfigure}{\textwidth}
        \centering
        \includegraphics{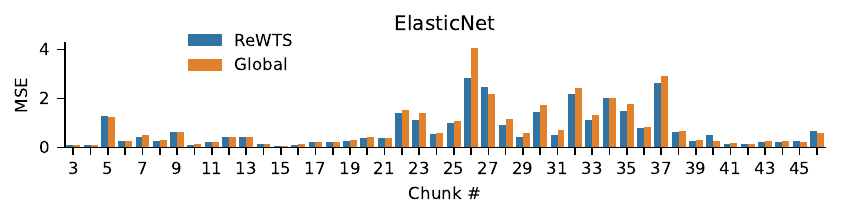}
        \caption{Elastic net model}
        \label{fig:elasticnet_veas}
    \end{subfigure}

    \begin{subfigure}{\textwidth}
        \centering
        \includegraphics{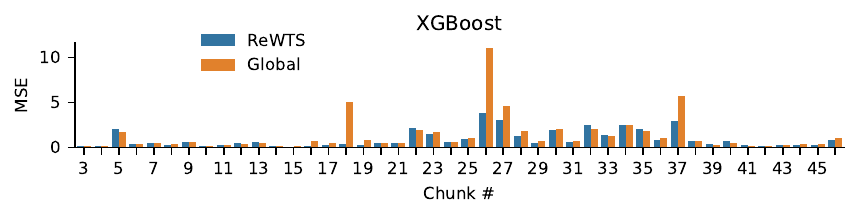}
        \caption{XGBoost model}
        \label{fig:xgb_veas}
    \end{subfigure}
    
    \begin{subfigure}{\textwidth}
        \centering
        \includegraphics{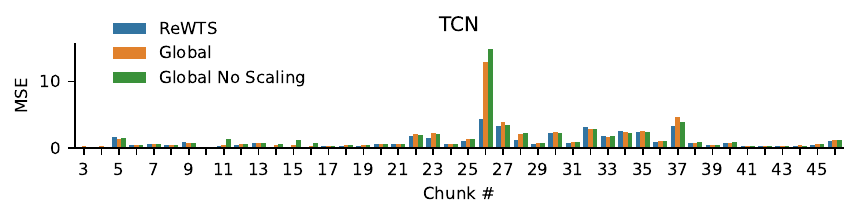}
        \caption{TCN model}
        \label{fig:TCN_veas}
    \end{subfigure}

    \begin{subfigure}{\textwidth}
        \centering
        \includegraphics{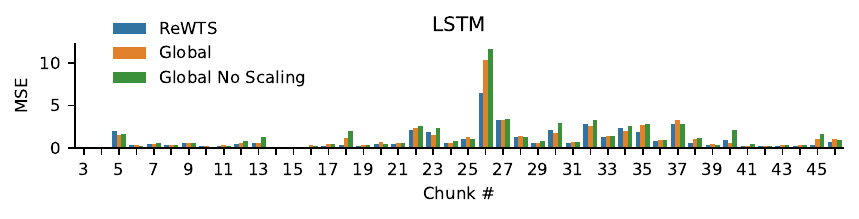}
        \caption{Vanilla LSTM model}
        \label{fig:LSTM_veas}
    \end{subfigure}
    \caption{Comparing the ReWTS ensemble model with the corresponding global model for the wastewater plant data for the model architectures elastic net, XGBoost, TCN and LSTM. The TCN and LSTM model were upscaled for each new chunk such as to match the total number of trainable parameters by the ReWTS ensemble model. For comparison, a global model without upscaling (Global No Scaling) is also presented.}
    \label{fig:veas_results}
\end{figure}

Figure \ref{fig:veas_results} shows the result when comparing the ReWTS ensemble model with the global model on the data from the wastewater treatment plant with chunk length of 2016 time points (corresponding to two weeks of data), and a look-back data length of 300 time points (corresponding to two days of data). 

In order to grasp the performance of each of the two procedures, we compute the loss function according to \eqref{loss} for \textit{each chunk}, a total of 44 chunks (first two chunks are omitted, as the ReWTS ensemble must necessarily have at least two models). We then compute the average loss, the average MSE, over all chunks for both the global model and the ReWTS ensemble model, separately for each model architecture. The numbers are given in Table \ref{tab:veas_result}.

\begin{table}[h!]
\centering
\begin{tabular}{ llrrrr } 
 \toprule
\multirow{4}{*}{Dataset} & & \multicolumn{4}{c}{Architecture} \\
\cmidrule(lr){3-6}
& & Elastic net & XGBoost & TCN & LSTM \\
\midrule
\multirow{3}{*}{Veas} & ReWTS & 0.73 & 0.92 & 1.01 & 0.99 \\  
&  Global & 0.81 & 1.15 & 1.33 & 1.16 \\ 
\cmidrule{2-6}
& \% Difference & 10.1 & 22.5 & 28.0 & 15.9 \\
\midrule
\multirow{3}{*}{Bergen} & ReWTS & 0.04 & 0.05 & 0.12 & 0.09 \\
& Global & 0.09 & 0.11 & 0.19 & 0.12 \\ 
\cmidrule{2-6}
& \% Difference & 74.6 & 65.9 & 48.4 & 33.9 \\
\bottomrule
\end{tabular}
\caption{The average MSE over all chunks of the data from the wastewater and drinking water treatment plants for the global model and ReWTS ensemble model, along with the percentage difference, for different model architectures.}
\label{tab:veas_result}
\end{table}

Consistently, for all model architectures, the global model is trailing the ReWTS ensemble model in terms of average MSE over the chunks. The reduction in average MSE is significant, and in the range from 10 to 28 \%, with the smallest reduction for the elastic net model, and the largest reduction for the TCN model. Interestingly, the linear model with regularization, elastic net, yields the best average MSE, both for the global model and the ReWTS ensemble model. This may be explained by the noisy measurements of the nitrate concentration where linear models can be more robust than more complex models that may put too much effort in modelling these random fluctuations.

\begin{figure}[t]
    \centering
    \begin{subfigure}{0.45\textwidth}
        \centering
        \includegraphics[width=\linewidth]{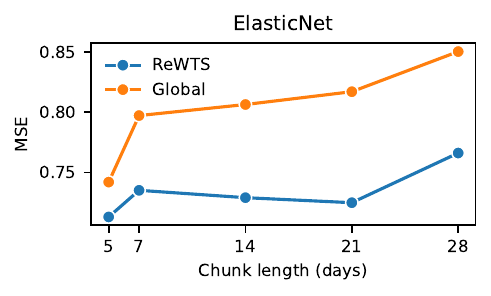}
        \caption{Elastic net model}
        \label{fig:elasticnet_veas_vary_chunk_length}
    \end{subfigure}
    \hspace{0.01 cm}
    \begin{subfigure}{0.45\textwidth}
        \centering
        \includegraphics[width=\linewidth]{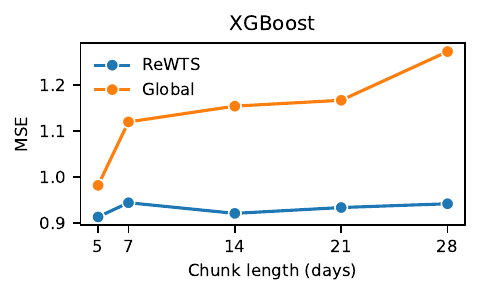}
        \caption{XGBoost model}
        \label{fig:xgb_veas_vary_chunk_length}
    \end{subfigure}
    \begin{subfigure}{0.45\textwidth}
        \centering
        \includegraphics[width=\linewidth]{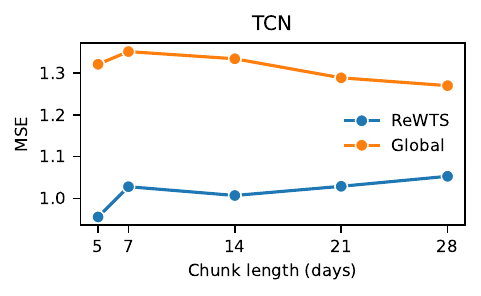}
        \caption{TCN model}
        \label{fig:TCN_veas_vary_chunk_length}
    \end{subfigure}
    \hspace{0.01 cm}
    \begin{subfigure}{0.45\textwidth}
        \centering
        \includegraphics[width=\linewidth]{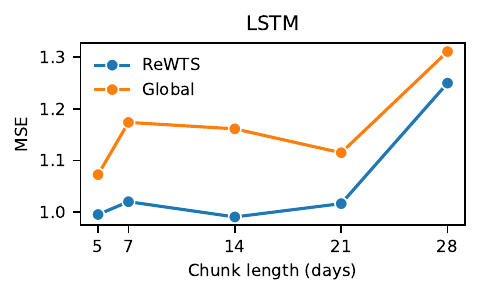}
        \caption{Vanilla LSTM model}
        \label{fig:LSTM_veas_vary_chunk_length}
    \end{subfigure}
    \caption{The average MSE, for different model architectures, of the ReWTS ensemble model for varying chunk length (with constant look-back length of 300 data points). Results when applied on the wastewater treatment data. The results are compared with the corresponding global model.}
    \label{fig:veas_results_vary_chunk_length}
\end{figure}

To further investigate the effect of the chunk length, we repeat the process above by keeping the look-back length constant at 300 data points, but vary the chunk length over 5, 7, 14, 21 and 28 days. The results are given in Figure \ref{fig:veas_results_vary_chunk_length} with chunk length on the x-axis and the average MSE over the chunks on the y-axis for the four model architectures.

Within this range of chunk lengths, the ReWTS ensemble model consistently outperforms the corresponding global model. The trend, while not consistent across all model architecture, is that the global model improves the more often it is updated (i.e. with smaller chunk lengths). There is also a trend that the ReWTS ensemble model improves the smaller the chunk length is.

\begin{figure}[h]
\centering
\includegraphics{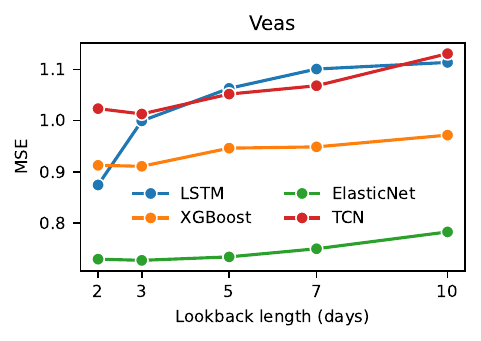}
\caption{The average MSE, for different model architectures, of the ReWTS ensemble model for varying look-back length (with constant chunk length equal to two weeks of data). Results are given for the wastewater treatment data.}
\label{fig:veas_results_vary_lookback_length}
\end{figure}

We do the same exercise to explore the sensitivity of the look-back length for the ReWTS ensemble model. We keep the chunk length constant to two weeks of data, and vary the look-back length over 2, 3, 5, 7 and 10 days of data. The results are given in Figure \ref{fig:veas_results_vary_lookback_length}.

The trend is clear, and consistent across all model architectures, that the smaller look-back length, the better performance with respect to the average MSE across the chunks.

In total, it is apparent that it is beneficial to keep both the chunk length, and the look-back length sufficiently small.

\subsubsection{Results on drinking water treatment plant}

\begin{figure}[t!]
\centering
\includegraphics{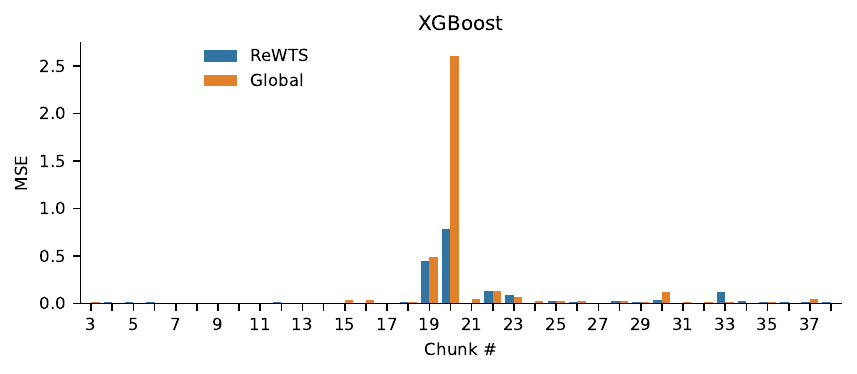}
\caption{Comparing ReWTS ensemble model with the corresponding global model for the water treatment plant data using the XGBoost model architecture.}
\label{fig:bergen_xgboost}
\end{figure}

Figure \ref{fig:bergen_xgboost} shows the result when comparing the ReWTS ensemble model with the global model, particularly for the XGBoost model, on the data from the water treatment plant with chunk length of 2016 time points (corresponding to two weeks of data), and a look-back data length of 300 time points (corresponding to two days of data). The corresponding plots for the elastic net, TCN and LSTM can be seen in Supplementary File.

 Table \ref{tab:veas_result} shows the average MSE for the different model architectures. The reduction in average MSE is significant, and in the range from 34 to 75 \%, with the smallest reduction for the LSTM model, and the largest reduction for the linear model.

The erroneous measurements described in Section \ref{sec:bergen} can be found between chunk 19 and 22. That's the reason why we see a sudden jump in MSE as a result of unseen, anomalous data. What is of particular interest is however what happens afterwards. Namely that the ReWTS ensemble model maintains a small MSE in future chunks, while the global model can have sudden spikes in the MSE (such as for chunk 30). When comparing the results with the wastewater treatment plant use case, we see that the MSE is in general smaller across all chunks. This is reasonable as the drinking water plant data has less complexity, and with less noisy measurements.
As for the wastewater treatment data, the average MSE of the ReWTS ensemble model is computed for varying chunk lengths and look-back lengths. The trends are mostly the same as for the wastewater treatment data, and for completeness the corresponding plots are given in Supplementary File.  

\subsubsection{Computational complexity of the ReWTS ensemble model}

We analyse the computational complexity of the ReWTS ensemble model and the global model. The computations can be separated in two separate tasks. The first task is the training of each model in its respective chunk. The second task is with respect to the forecasting. Note that for the ReWTS ensemble model, according to the procedures in \ref{alg:chunk_model_forecasting}, forecasts need to be evaluated within the look-back data to compute the weights, before eventually providing the forecasts. Meanwhile, the global model can at once provide the forecasts.

\begin{figure}[t!]
    \centering
    \begin{subfigure}{\textwidth}
        \centering
        \includegraphics{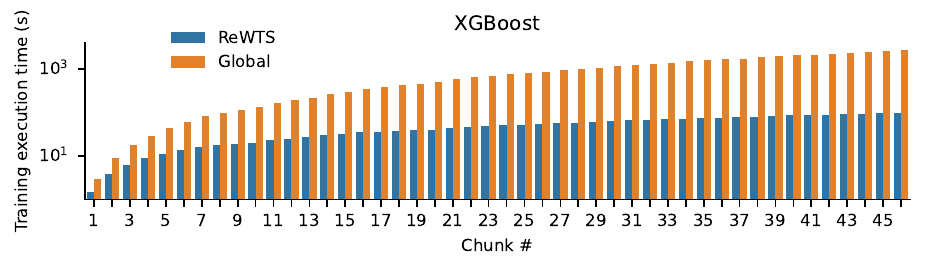}
        \caption{Training}
        \label{fig:execution_time:training}
    \end{subfigure}
    
    \begin{subfigure}{\textwidth}
        \centering
        \includegraphics{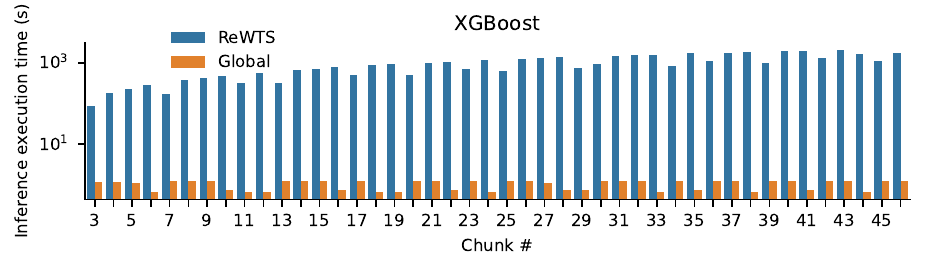}
        \caption{Forecasting}
        \label{fig:execution_time:inference}
    \end{subfigure}
    \caption{Execution time of training (a) and forecasting (b) for the ReWTS ensemble model using the XGBoost architecture.}
    \label{fig:execution_time}
\end{figure}

Figure \ref{fig:execution_time} illustrates the computational time required for both the training and forecasting procedures when employing the ReWTS ensemble model and the global model on wastewater treatment data, using the XGBoost architecture. In Figure \ref{fig:execution_time:training}, we depict the total training time after completing each chunk, comparing the ReWTS ensemble model and the global model with a logarithmic scale on the y-axis. The ReWTS ensemble model updates by training a new chunk model with a fixed number of data points, defined by the chunk length. In contrast, the global model updates by retraining on an increased amount of data points equal to the chunk length. Consequently, the ReWTS ensemble model requires less computational time in this regard, and the difference in training time between the two methods amplifies over time.

Figure \ref{fig:execution_time:inference} shows the computational time required for the forecasting process, in each chunk respectively, for the global model and the ReWTS ensemble model. The computational time is significantly larger for the ReWTS model than the global model, and the difference increases as more chunk models are added to the ReWTS ensemble. This is because the number of forecasts executed on the look-back data, as part of the weight fitting procedure, increases with the number of chunk models. The global model on the other hand is not relying on any look-back data, and can provide the forecasts immediately at prediction time.

\section{Discussion and conclusion} \label{sec:Discussions}

The ReWTS ensemble model introduced in this paper is simple, and easy to implement. For the simulated data, where we know exactly where the different data generating processes are, we show that the ReWTS ensemble model outperforms a global model trained with all data in one go. The same applies for noisy real process industry data. This conclusion pertains to two distinct datasets from different processes at different plants (wastewater treatment and drinking water treatment). It underscores a noteworthy reduction in the mean squared forecasting error, ranging from 10\% to 70\%, across diverse model architectures. Importantly, the ReWTS ensemble model demonstrates increased resilience to outliers. In the case of real process industry data, it is not the case that each data chunk represents a static data generating process. However, the chunk may include characteristics that are unique, or available only in a few other chunks. Hence, each chunk model can specialize within a distinct time interval with restricted heterogeneity in the dynamics. We regard this is as the most likely reason for the ReWTS model to outperform the global model.

By varying the chunk length and look-back length, we see that for both parameters it is beneficial to keep the length sufficiently small. Regarding the chunk length, this is intuitive in the case where a smaller chunk length provides more distinct, and less diverse characteristics in the chunk. The smaller look-back length, the more will the weight assignment in the ReWTS ensemble model be based on the most recent data. This can be an advantage in cases where the dynamics change rapidly. Domain knowledge about characteristic timescales in the system will often guide these choices. We leave the exploration of disjoint, overlapping or different length chunks to future work.

An additional benefit of the ReWTS ensemble model, as opposed to the global model, lies in its enhanced capacity for interpretability and reliability analyses, namely by examining the weights of the different chunk models. One can assess the credibility of the forecasts by analyzing the dynamics within the chunks that corresponds to the chunk models that were assigned the largest weights.

When comparing the ReWTS ensemble with a corresponding global model, the number of trainable parameters were upscaled for the neural network architectures when retraining the global model, while not for the linear and XGBoost model. A procedure for upscaling the global XGBoost model was implemented by making sure the maximum number of regression trees from the global model would match the maximum number of regression trees from the corresponding ReWTS ensemble model. This did not lead to any significant changes in performance of the XGBoost model.

As discussed in Section \ref{sec:related}, there exist several proposals on how to discard under-performing chunk-models. By the construction of the weight assignment in ReWTS, the weights assigned to each model can give an indication of how important it is. For instance a model that is often assigned a small weight, and with small variation, can be considered as irrelevant in most cases. In this case, one could consider to discard it in future weight updates. We leave this as future research.

In this work we have restricted ourselves to the mean squared error both for assigning the weights in the ReWTS ensemble model, and to evaluate the performance of the forecasting models. In principle, any other similarity measure can be applied to decide the weights in the ReWTS ensemble model. Note that the metric can be different between the model training and the weight assignment. This allows for reuse of existing models of particular phenomena in the system to be combined with purely data driven models.

We have seen that the weight assignment procedure of the ReWTS ensemble model is computationally heavy and depends on the look-back length. A possible procedure to reduce the computation time is to only use a subset of the look-back data, e.g. by using a stride larger than one.

For the wastewater data set, the period with known error measurements were manually modified with linear interpolation. Further, for the drinking water treatment plant, the known region of error measurements were ignored, and the raw data was used during training. For a fully-automated forecasting model, the ideal use-case would be to detect anomalies, such as error measurements, and do the appropriate corrections on the fly instead. How this should be done in practice is not straightforward, and we leave this for future research.

A tempting alternative to the ReWTS ensemble model would be to train several forecasting models with each forecasting model trained on input sequences of similar dynamics, and then construct an ensemble model based on this. With respect to the wastewater treatment data, this means for instance to have one model trained to forecast future nitrate concentrations based on input sequences of small and stable (low variance) nitrate concentrations. Yet another model can be trained with input sequences where the nitrate concentration is rapidly increasing. To group the input sequences into different classes or cluster will require some assumptions or decisions. With this alternative grouping of models, at prediction time, the models are weighed based on how much the input sequence resembles the individual clusters, therefore eliminating the usage of look-back data. This concept is in accordance with the initial idea behind ReWTS ensemble modelling, namely that each model can specialize in particular patterns of the data. However, for this procedure to adapt to new patterns in the data, it is necessary to be able to assign new clusters of sequences and corresponding new fitted forecasting models. How and when to assign new clusters is left as future research.

\section{Acknowledgments}
We are grateful for the seamless sharing of data from Veas wastewater treatment plant and Bergen Vann through the drinking water treatment plant in Bergen, and for the fruitful and constructive co-operation with the great colleagues at the two respective treatment plants.
We thank our colleague Helga Holmestad at SINTEF for her contribution regarding discussions of conceptual idea as well as reviewing. We express our gratitude to our colleague, Bård Myhre, for engaging in insightful discussions, solid project management, and serving as a valuable bridge between our team at SINTEF and the water treatment plants.

This research has been funded by The Research Council of Norway, grants 309517 (INVAPRO), 294544 (TAPI), as well as by the EU HORIZON-IA program, grant 101135932 (FAITH).

\section{Supplementary information}
The Supplementary File referred to in this article is available as an ancillary file in the published arxiv version.

\section{Source code, data Availability and access}

The source code used for providing the results in this work is openly available at \url{https://github.com/SINTEF/rewts}. The results of the simulation data described in Section \ref{sec:sim_data_description} can be reproduced as detailed in the README file. Also described in the README file is how to deploy the ReWTS ensemble model on your own data set. 

The drinking water treatment data used in this project require handling with utmost sensitivity, given their direct implications for public health, safety, and privacy. Access to the data platforms from which this project has sourced its information may be granted upon formal request from qualified individuals or organizations, specifically for research purposes or with expediency justified by the data providers. Requests for access should adhere to established ethical and legal considerations. Interested parties are invited to contact Bergen Vann for further details, and to initiate the request process.

Comprehensive understanding and appropriate utilization of wastewater treatment data necessitate additional knowledge about Veas. Veas ensures both this knowledge and adherence to ethical and legal guidelines by sharing data through agreement-based methods. If you wish to access data for development or research purposes, feel free to reach out to Veas.

\printbibliography

\end{document}